# Stable Diffusion for Data Augmentation in COCO and Weed Datasets

Boyang Deng
Department of Biosystems & Agricultural Engineering, Michigan State University,
East Lansing, MI 48824, Correspondence: dengboy1@msu.edu

## Abstract

Generative models have increasingly impacted various tasks, from computer vision to interior design and beyond. Stable Diffusion, a powerful diffusion model, enables the creation of high-resolution images with intricate details from text prompts or reference images. An intriguing challenge lies in improving performance for small datasets with image-sparse categories. This study explores the effectiveness of Stable Diffusion by evaluating seven common categories and three widespread weed species. Synthetic images were generated using three Stable Diffusion-based techniques: Image-to-Image Translation, DreamBooth, and ControlNet, each with distinct focuses. Classification and detection tasks were then performed on these synthetic images, and their performance was compared to models trained on original images. Promising results were achieved for certain classes, demonstrating the potential of Stable Diffusion in enhancing image-sparse datasets. This foundational study may accelerate the adaptation of diffusion models across various domains.

## Introduction

Diffusion models are probabilistic generative models that learn data distributions by progressively denoising variables sampled from a Gaussian distribution (Yang et al., 2023). They decompose the image formation process into a sequential application of denoising autoencoders. However, traditional diffusion models (DMs) operate directly in pixel space, making optimization and inference computationally expensive and time-consuming (Croitoru et al., 2023). Stable Diffusion (SD, Rombach et al., 2022) introduced Latent Diffusion Models (LDMs), which apply diffusion processes in the latent space of powerful pre-trained autoencoders. This approach enables stable and controllable information diffusion through neural network layers, facilitating high-fidelity image generation. Additionally, Stable Diffusion significantly reduces computational demands compared to pixel-based DMs, marking a major advancement in image synthesis.

This model has led to a wide range of derivatives, including ControlNet, Dream Booth, and LoRA , as well as various applications such as image inpainting, class-conditional image synthesis, and text-to-image generation. It has also inspired commercial products like Draw Things, DreamStudio, and NovelAI.

Stable Diffusion has the capability to generate realistic, high-fidelity images that align with predefined layouts. Notably, additional guiding mechanisms, such as ControlNet, can be integrated into diffusion models to refine the image generation process without requiring retraining.

ControlNet (Zhang et al., 2023) is an end-to-end neural network architecture that enhances the controllability of Stable Diffusion by incorporating additional conditions. It leverages a large pre-trained model as its backbone while freezing its parameters and creating a trainable copy of the encoding layers, which are linked to the original model. To prevent disruptive noise from affecting the deep features of the diffusion model during the initial training phase, the trainable copy is initialized with zero weights. This design enables ControlNet to maintain the quality and capabilities of the large model while effectively learning diverse conditional controls. ControlNet can achieve remarkable results across a wide range of applications. For instance, a new image of a person with a specific gesture can be generated by specifying a desired human pose using an existing full-body portrait or an online pose generator like MagicPoser, which manipulates arbitrary poses in a 3D person model. Pose detection networks, such as OpenPose, are then used to detect key nodes of the pose in the reference image. By combining the detected pose with a text prompt (e.g., "an awesome man in a suit"), a new image is generated that adheres to both the pose and the prompt. This pose-guided person generation process has significant potential to improve person category generation, as discussed in this study. Figure 2 illustrates this pipeline.

Recent text-to-image models, trained on large collections of image-caption pairs, can generate high-quality and diverse images from text prompts (Ramesh et al., 2022). However, they struggle to accurately mimic the appearance of specific subjects in reference sets and create novel renditions of these subjects in different contexts, as the expressiveness of their output domain is limited (Ruiz et al., 2023). Even with detailed textual descriptions, these models often produce instances with varying appearances. Existing models are not designed to reconstruct the exact appearance of given subjects but instead generate variations of the content. However, Stable Diffusion, when combined with DreamBooth or ControlNet, provides new solutions to these challenges, which will be discussed in the following sections.

Weed detection is a critical step in enabling precise weed management with automatic robots. The Garford Robocrop InRow Weeder (Tillett et al., 2008) featured a disk-like structure mounted on a depth control wheel designed to remove weeds within crop rows while avoiding crop damage. Its robot was guided by a centrally mounted camera, and a Kalman filter tracking algorithm (Bar-Shalom et al., 2011) was used to predict plant positions. Underwood et al. (2015) introduced a mobile robot utilizing a range of sensors, including LiDAR, hyperspectral imagers, and GPS, to detect and locate weeds. The AgBotII (Bawden et al., 2017), a vehicle-like machine, employed a downward-facing camera and a lighting module for weed detection and classification. The Deepfield Bonirob (Lottes et al., 2017) was deployed to collect a weed dataset with seven species in a sugar beet field, utilizing multiple sensors to capture multimodal data such as RGB, NIR (near-infrared), and depth images from a Kinect v2 sensor, point clouds from laser scanners, and robot-related data like wheel odometry. Gai et al. (2020) monitored broccoli and lettuce with their robotic vision system, capturing color and depth images using a Kinect v2 sensor mounted on a mobile platform. These robots facilitate the collection of large-scale weed image datasets, providing ample data for developing detection models.

The shortage of large-scale datasets in specific fields, such as weed management, poses a significant challenge to the development of advanced data-driven models that require vast amounts of data for training. This study aims to explore the potential of Stable Diffusion-based techniques to generate synthetic images that can complement realistic images in real-world applications. Our objectives were threefold:

1. We evaluated the effects of the image-to-image translation of Stable Diffusion on object detection tasks based on the COCO and CottonWeedDet12 datasets.
2. We fine-tuned the Stable Diffusion model through DreamBooth to generate class-specific images and assessed whether these images could enhance performance in a classification task.
3. Using the ControlNet-added Stable Diffusion, we generated human images of different ages and poses, then assessed their utility in a detection task focused on the person class.

# Method

## 2.1 Dataset

Two datasets, COCO (Lin et al., 2014) and CottonWeedDet12 (Dang et al., 2023), were used to generate synthetic images. The COCO dataset offers per-instance annotations, including bounding boxes and segmentations, to assist with precise object localization. It contains around 2.5 million labeled instances across 328,000 images, with 91 distinguishable categories. Of these, the 80 categories from the 2014 release are the most used in research. This study focused on seven categories: person, dog, cow, train, car, motorbike, chair, sofa, and bottle. These categories represent various everyday objects with diverse attributes and characteristics. For example, the "person" category includes a wide range of poses (e.g., standing, riding a horse, dancing) and backgrounds (e.g., city stations, indoor, natural fields), making it challenging for detection models due to the large number of instances in COCO. In contrast, the "train" category tends to perform better in detection models because of its more rigid shapes and consistent environments.

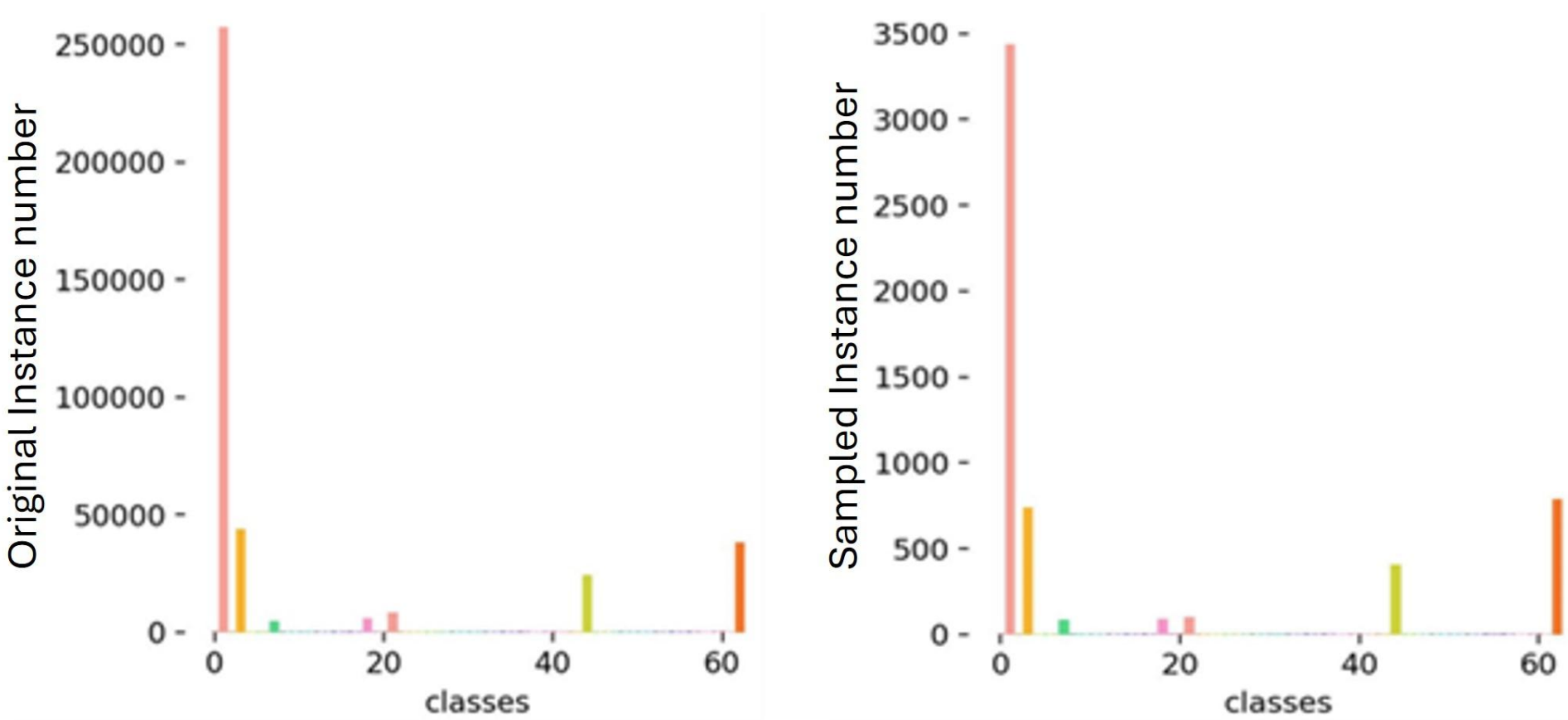

Figure 1. The instance distribution across the target classes of COCO. On the left are all the instances of each class, while on the right are the numbers for the sampled instances. The target classes are as follows: 1: person, 3: car, 7: train, 18: dog, 21: cow, 44: bottle, 62: chair.

CottonWeedDet12 consists of 5,648 images representing 12 weed species, collected for weed detection. For this study, three weed species (i.e., Purslane, Ragweed, and Palmer Amaranth) were selected as targets, as they are common in Michigan and widespread across the U.S. Dang et al. (2023) performed comprehensive tests of 18 YOLO models on this dataset, demonstrating promising results for YOLO models on weed images captured in various fields. However, there is still room for improvement, particularly for specific weed species like those chosen in this study. The dataset includes 111 images of Palmer Amaranth, 191 images of Purslane, and 149 images of Ragweed. Figure 2 shows the distribution of selected wed species.

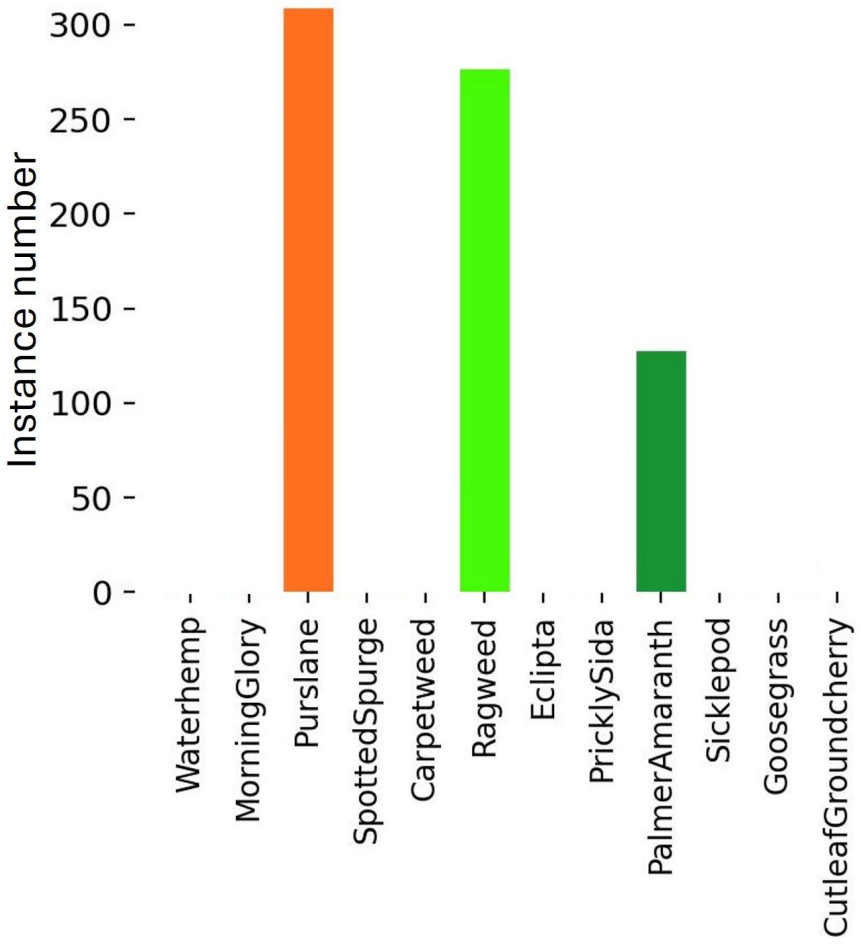

Figure 2. The instance distribution across the target classes of COCO. Only three of the 12 weed species were selected for experiments. 3: Purslane, 6: Ragweed, 9: PalmerAmaranth.

Compared to the more general object categories in the COCO dataset, CottonWeedDet12 is tailored to weed images and is relatively small in scale. This makes it a step forward from general object detection to specialized tasks involving small or medium-scale datasets focused on specific subjects in various fields (e.g., molecule detection, ship detection in satellite imagery, or shelf beverage detection). Figure 3 shows the examples of target weed species.

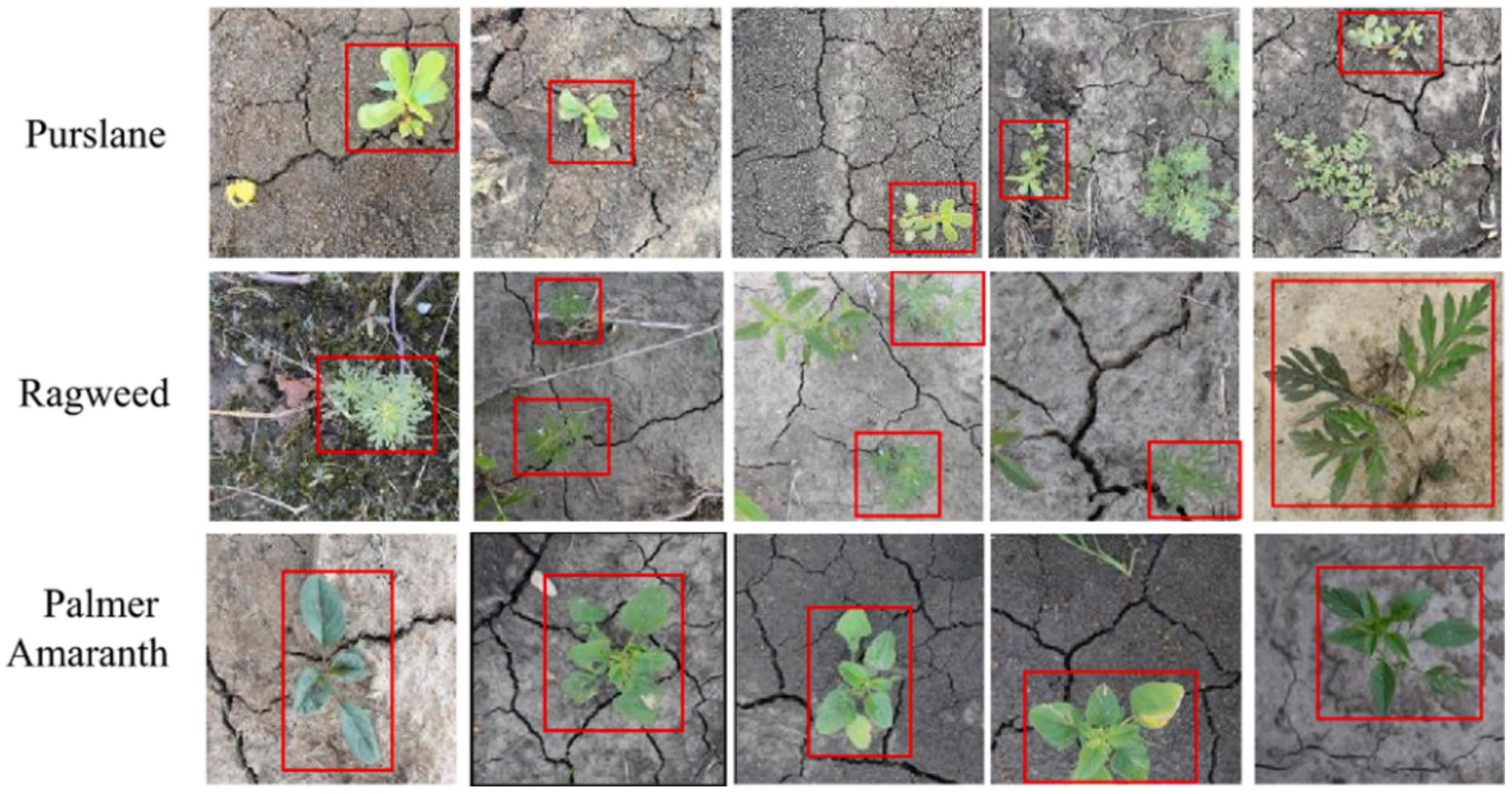

Figure 3. Examples of the three weed species.

## 2.2 Image to image generation

Stable Diffusion's image-to-image translation, available in the Stable Diffusion Wambui application, generates synthetic images based on a text prompt and an initial image. The text prompt specifies the desired content, such as style, pose, or background changes, while the initial image serves as a base for easier learning and constrains the model's output. Key parameters include the strength parameter, which adjusts the noise added to the initial image for consistency, and the CFG, which controls the influence of the text prompt, with larger values having a stronger impact.

Figure 4 compares the generated images to the originals. The generated images capture the key features of the target classes well, though some details, like human faces and the black markings on Palmer Amaranth leaves, are less accurate. Image numbers from COCO categories and weed species in CottonWeedDet12 were down sampled to 1/100, and each image generated four counterparts via image-to-image translation, forming a new dataset with both realistic and synthetic images. These new datasets were trained on the YOLOv8-large model, and the results will be compared to YOLOv8 models trained on only 1/100 realistic images or all the original images from the seven classes.

The CLIP interrogator helped generate prompts, with a positive prompt of "great details, more variations, high fidelity, realistic style" and a negative prompt of "blur, low quality, unpleasant, stylistic genre." The CFG Scale was set to 7.5, and denoising strength to 0.35 to preserve similarity to the original image and minimize annotation errors. Twenty sampling steps were used, with all images sized at 800x800 pixels. The "resize" strategy adjusted original images, and Euler sampling was chosen for efficiency. Stable Diffusion v1.5 was used with these settings.

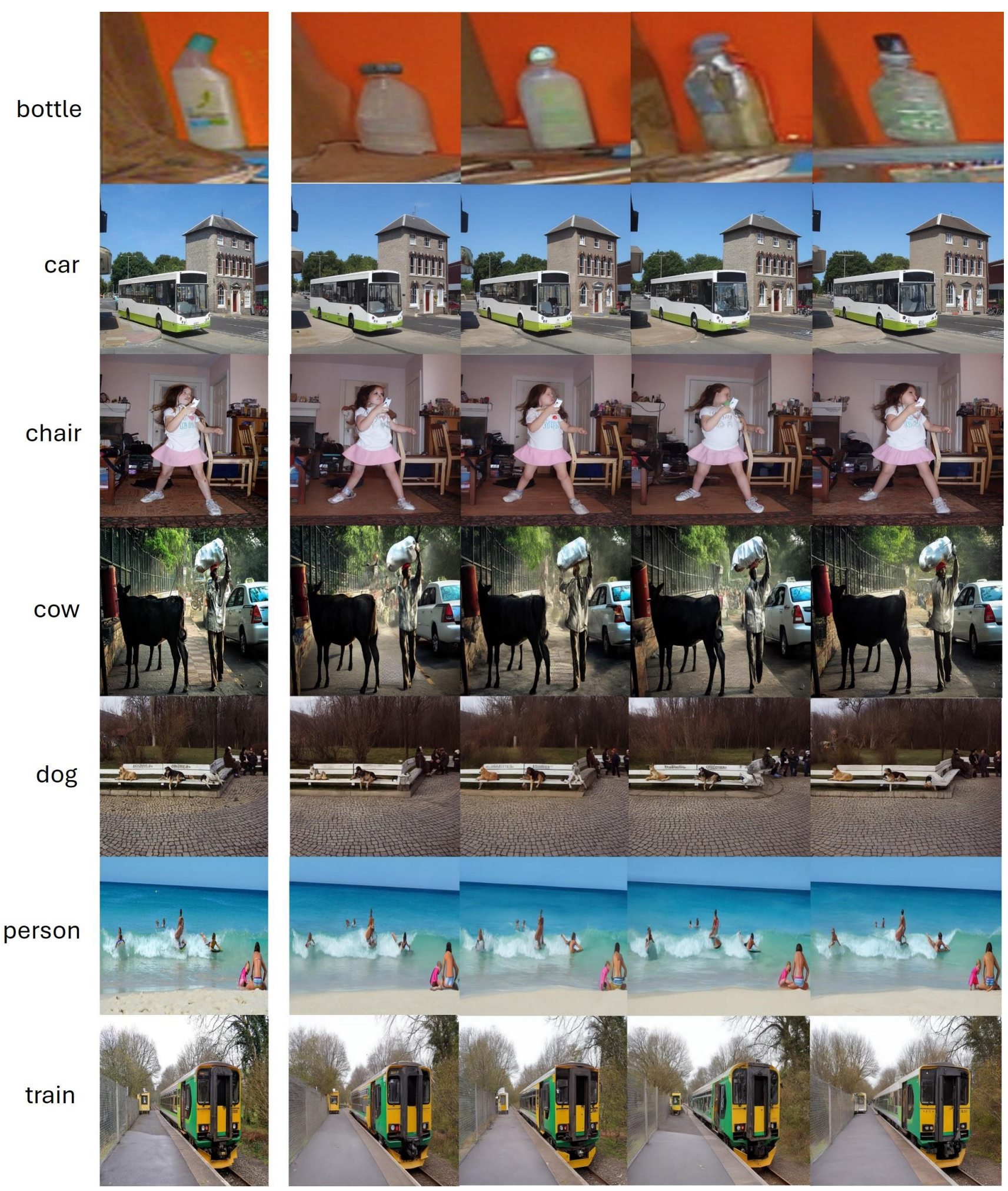

(a) The left column shows the original images from seven categories of COCO, while the right column displays the generated examples.

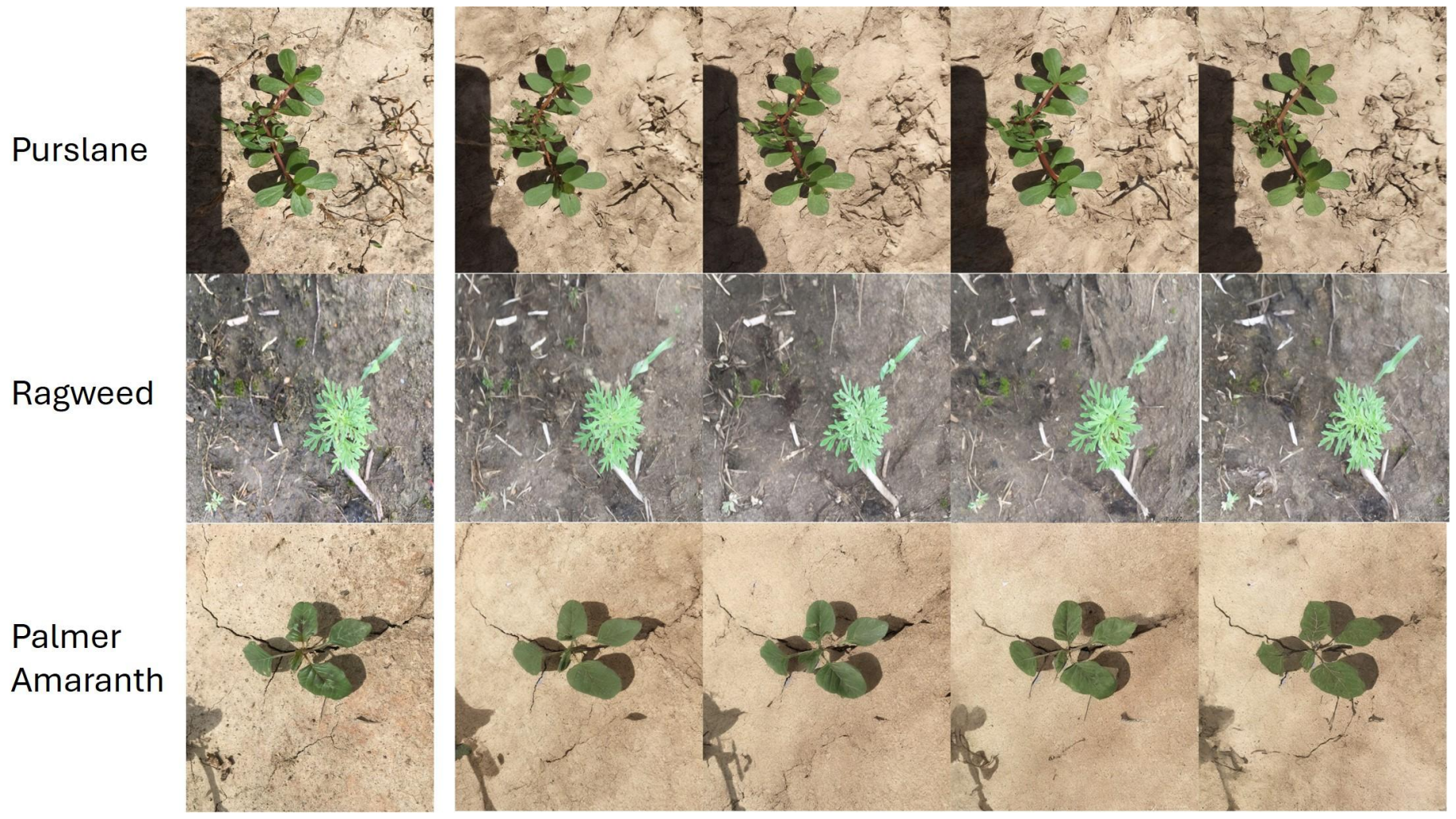

(b) The left column shows the original images from three categories of CottonWeedDet12, while the right column displays the generated examples.

Figure 4. Examples derived from image-to-image translation. The left is examples of the original images, while the right are the four columns of synthetic images.

### 2.3 DreamBooth-based model finetuning

DreamBooth (Ruiz et al., 2023), published in 2022 by a Google research team, is a method for efficiently fine-tuning Stable Diffusion to inject a custom object into the model. By using only 3-5 reference images and text prompts, DreamBooth allows for the generation of a wide range of images of a specific subject in various contexts. The name comes from the original description by the team, likening it to a photo booth. Once the subject is captured, it can be synthesized anywhere your imagination takes you.

DreamBooth personalizes text-to-image models by synthesizing specific subjects in new contexts. It expands the model's language-vision identifiers, associating new words with the user's desired subjects through a class-specific prior preservation loss. After embedding these identifiers, the model can generate novel, photorealistic images of the subject in various scenes while preserving key features. This technique fine-tunes a pre-trained diffusion model like Stable Diffusion using rare token identifiers for the subject.

The unique identifier in the prompt should be rare to avoid overlap with frequently used words the model has memorized. DreamBooth's original paper used "[V] dog" as an example to represent a specific dog species. In this study, each class is fine-tuned separately by embedding "[V]" into the class-specific prompt for the COCO dataset (e.g., "photo of [V] person"), compared to the reference prompt "photo of person." This identifier is intended to link strongly with the reference images for each class. For CottonWeedDet12, since weed species names are rare in Stable Diffusion, we use prompts like "photo of [weed species name] weed" (e.g., "photo of Purslane weed") and "photo of weed" as the reference for all weed species.

DreamBooth was used to generate new images for the seven categories in COCO and the three weed species in the CottonWeedDet12 dataset, following the same synthetic image generation protocols as Image-to-image generation for a fair comparison. Ten reference images of each category were used to fine-tune the original Stable Diffusion model for generating the corresponding categories. Although DreamBooth allows for more varied prompts (e.g., background, viewpoints, and styles), the same positive and negative prompts as Image-to-image generation were applied across all image generations.

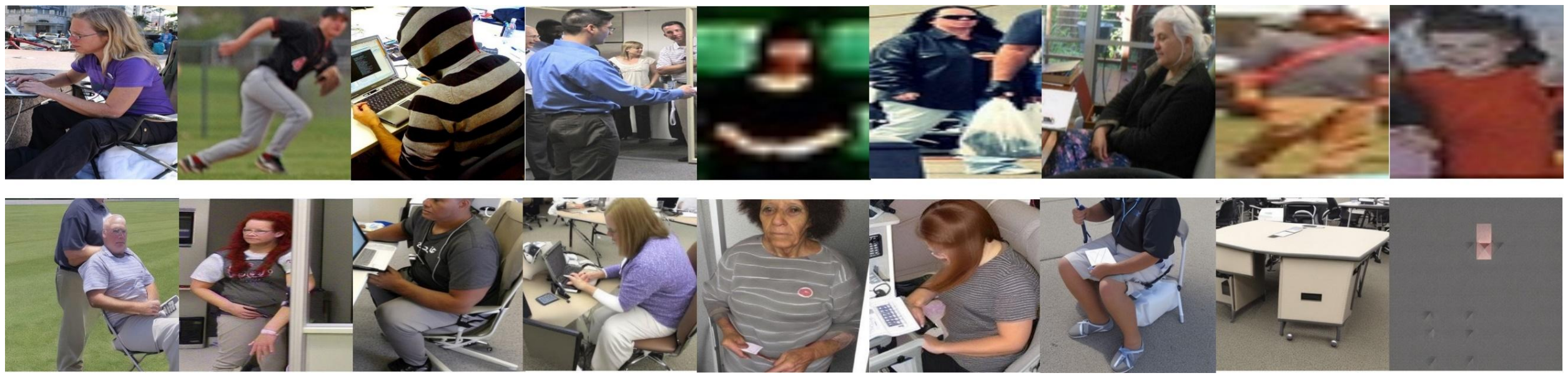

Figure 5. The first row is the reference images of the “person” class for Dreambooth. The second row shows the Dreambooth-generated examples. The last two images in the second row show the failed cases.

At this stage, it takes only 10 minutes to fine-tune a model for each class. However, failures are observed in some categories (e.g., person, car), while others (e.g., chair, cow) generate high-quality images. For the "person" class, 20%-30% of generated images fail to show any human or produce unrecognizable content, which will be removed before classification. Initially, 150 images were generated for each category, and 50 were manually deleted to build the training set. Interestingly, the model can generate out-of-distribution images, some with better fidelity than the reference images, which may benefit from Stable Diffusion’s capabilities. Additionally, some images feature partial human bodies, which could complement the original full-body images. Stable diffusion has learned vast information from text and images, enabling high-quality image generation of general objects, such as those in the COCO dataset. Comparing synthetic images from non-finetuned and Dreambooth-finetuned models, this study also evaluates classification performance with both image types.

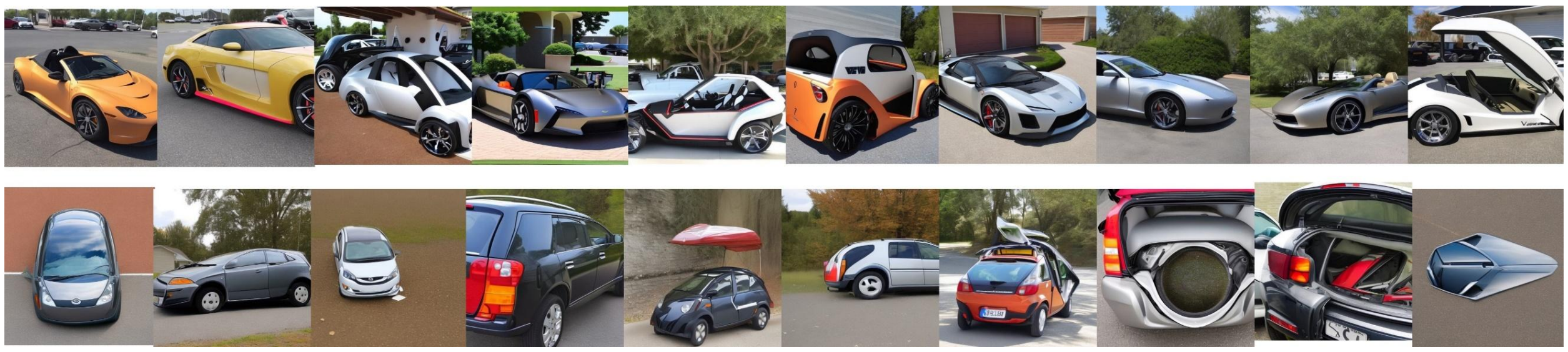

Figure 6. The images from the original stable diffusion model (top) and the Dreambooth-finetuned stable diffusion model (bottom).

Due to the complexity of annotating generated images, bounding box-cropped images from each class were used to finetune the original Stable Diffusion network through Dreambooth. This process created a new COCO-BBox dataset, where each class's bounding box formed sub-region images. After fine-tuning, 100 synthetic images were generated for each category. A classification task then assessed the validity of these generated images by training a model on the 100 synthetic images. This model was compared to one trained with just ten images per category (used for Dreambooth) and another trained with 100 original images (including the ten used for Dreambooth). The COCO-BBox dataset includes 700 training images, while the CottonWeedDet12 dataset has 300, with consistent test and validation sets across all experiments.

## 2.4 ControlNet-based synthetic images

Pose-based variations of the ControlNet can generate the category of person in the COCO dataset. the pose generation website (https://webapp.magicposer.com/) was leveraged to generate various poses for realistic personal images. The bounding box of each pose in the pose image was utilized as the annotation for the synthetic images generated from this pose image. 5 poses were designed to cover the daily gestures of humans, observed from the person class of COCO, and 800 images were generated from each pose. Notably, there are some bounding boxes of "person" that only contain the upper part of even the head of the target person, while there are rare bounding boxes that only contain the below body of the human without the head in COCO, which follows the intuitive of human annotation that head is the most important part to identify a human. Figure 7 shows the various poses of "person" in COCO.

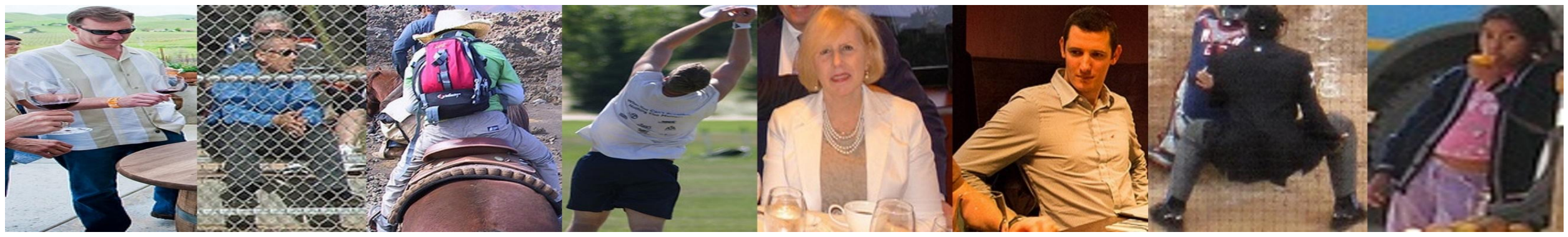

Figure 7. Examples of the "person" class in COCO, with various poses and viewports.

This study found that excluding "realistic style" from the prompt for human generation in ControlNet produced more realistic images, while including it made them appear more cartoonish. "Realistic style" was included for kids and excluded for other groups. The positive prompt for kids was "a [kid], [boy | girl], great details, more variations, high fidelity," and for others, it was "a [young adult | middle-aged | elder], [man | woman], great details, more variations, high fidelity." The negative prompt remained the same. Four age-gender groups (kids, young adults, middle-aged, and elders) were generated with 100 images each. ControlNet used a 3D model with the pose_full preprocessor and generated images at 512x512 resolution to save computation resources.

Figure 8 displays the pose images alongside the generated ones, revealing interesting phenomena. For instance, the kids' images show hallucinations, with facial regions often imperfect, but the overall quality still effectively represents humans of various ages and poses. Interestingly, elderly individuals tend to appear more cartoonish, even when the prompt provides no clear indication of this.

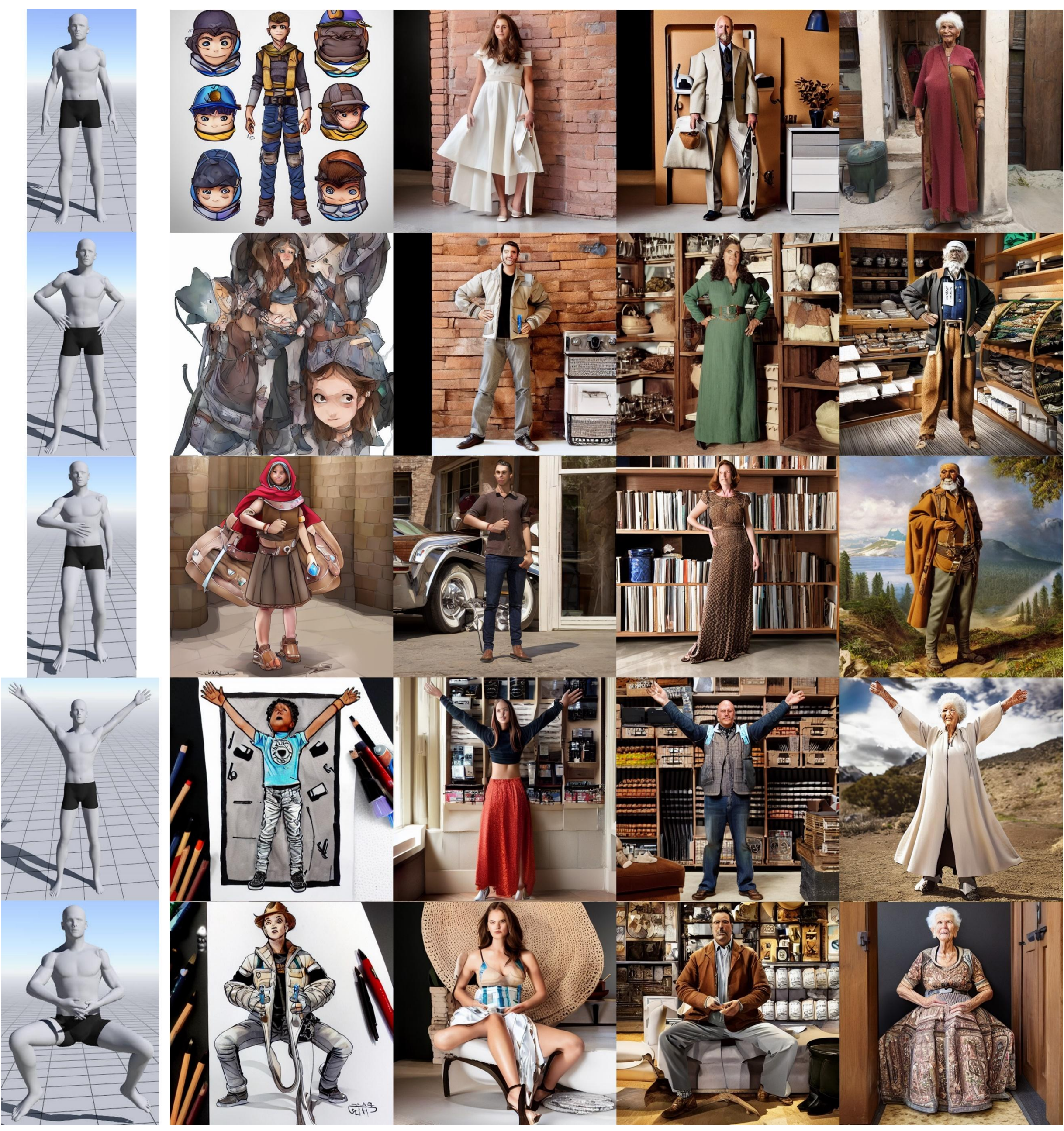

Figure 8. the examples of the generated images of all four ages with the five poses.

## 2.5 YOLOv8 detection model

Deep learning-based object detectors typically consist of three primary components, i.e., the backbone, the neck, and the head (Terven & Cordova-Esparza, 2023). The backbone extracts features from input images, typically using a convolutional neural network trained on tasks like ImageNet. The neck refines and aggregates these features, linking the backbone to the head. The head generates predictions based on these features, usually through task-specific subnetworks like classification, location, and segmentation.

YOLO series models are popular in real-time weed detection, benefiting from their high inference speed and decent accuracy (Dang et al., 2023). YOLOv8, released in January 2023 by Ultralytics (the creators of YOLOv5), introduced several enhancements, including the innovative CSPDarknet53 backbone, distribution focal loss (DFL) for class imbalance, and a one-stage task-aligned assigner for aligning classification and location tasks. These improvements enable the YOLOv8-large model to offer advanced detection performance with high inference speed, making it ideal for weed detection.

### 2.6 Classification model

The torchvision model of vit_l_16, using the IMAGENET1K_SWAG_E2E_V1 version, achieved 88.064% accuracy at top-1 and 98.512% at top-5 on the ImageNet-1K dataset, and was used for our classification tasks. All classification experiments were conducted at a 512x512 resolution, with image preprocessing performed using the Python library of augmentation 1.3.1.

## Experiments

### 3.1 image to image generation

Four experiments were conducted for detection tasks on each dataset. The first evaluated model performance on all training images of the classes. The second tested performance on sample images used to generate synthetic images. The third assessed performance on the synthetic images, and the fourth combined both sampled original and generated synthetic images. All experiments used 12 epochs. The first experiment took about 24 hours, using a large-scale training set with 80,354 images and 381,013 bounding boxes. The latter experiments required 3 hours each. The second dataset had 1,059 images with 5,626 bounding boxes, and the third dataset had 4,236 images with 22,504 bounding boxes. The fourth experiment merged the second and third datasets. YOLOv8-large was used for weed detection models, assessed on the test set using mean average precision (mAP@50). The experiments were conducted with PyTorch (v1.13.1) on the NVIDIA RTX A6000.

Table 1. The results of the COCO dataset.

| COCO (mAP@50) | all | person | car | train | dog | cow | bottle | chair |
|---|---|---|---|---|---|---|---|---|
| All images | **0.808** | **0.863** | **0.78** | **0.94** | **0.881** | **0.89** | **0.682** | **0.621** |
| 1% images | 0.781 | 0.842 | 0.756 | 0.922 | 0.857 | 0.858 | 0.642 | 0.59 |
| Synthetic images (4%) | 0.725 | 0.821 | 0.711 | 0.886 | 0.774 | 0.697 | 0.596 | 0.59 |
| 1% Original+Synthetic (4%) | 0.759 | 0.836 | 0.739 | 0.91 | 0.789 | 0.815 | 0.642 | 0.582 |

Table 1 presents the results based on COCO dataset. "All images" refers to all images from the seven COCO classes. "1% images" are subsampled images used to generate synthetic images, while "Synthetic images (4%)" refers to the synthetic images, which make up 4% of the total images in the seven classes.

The last row combines 1% original and 4% synthetic images. Notably, the training set comprising 1% original and 4% synthetic images performed better than using only synthetic images, but worse than using just 1% original images. This suggests that directly adding synthetic images to the training set may negatively impact the training process. Overall, the model trained solely on original images achieved the best results.

Table 2. The results of the CottonWeedDet12 dataset.

| CottonWeedDet12 (mAP@50) | all | Purslane | Ragweed | PalmerAmaranth |
|---|---|---|---|---|
| Original images | **0.846** | 0.837 | 0.961 | **0.739** |
| Synthetic images | 0.737 | 0.766 | 0.968 | 0.477 |
| Original+Synthetic images | 0.791 | **0.874** | **0.974** | 0.524 |

Table 2 shows the detection results for the CottonWeedDet12 dataset. While the original-only training set achieved the best overall performance, combining original and synthetic images improved detection for Purslane and Ragweed, highlighting the effectiveness of synthetic images. However, performance for PalmerAmaranth dropped significantly when synthetic images were included, possibly due to the generated images not closely resembling realistic PalmerAmaranth features. This suggests that more universal methods for generating images across all classes are needed, and stable diffusion models could help bridge these gaps in future research.

### 3.2 DreamBooth-finetuned results

For COCO and CottonWeedDet12 datasets, three classification experiments were conducted. The first evaluated the model performance on all bounding box images of the classes in the training set. The second tested performance on ten bounding box images from each class, which were used to fine-tune the Stable Diffusion model via DreamBooth. The third assessed performance on the synthetic bounding box images. Table 3 shows the DreamBooth-finetuned results based on COCO dataset.

Table 3. The DreamBooth-finetuned results of COCO.

| COCO (mAP@50) | all | person | car | train | dog | cow | bottle | chair |
|---|---|---|---|---|---|---|---|---|
| 1% images | **0.83** | **0.899** | **0.788** | 0.9 | 0.766 | 0.777 | 0.608 | 0.597 |
| Reference images | 0.31 | 0.232 | 0.469 | 0.9 | 0.156 | 0.79 | 0.59 | 0.301 |
| Synthetic images (4%) | 0.635 | 0.648 | 0.736 | 0.937 | 0.743 | 0.723 | 0.562 | 0.425 |
| Reference+Synthetic images | 0.608 | 0.589 | 0.737 | **0.947** | **0.821** | **0.849** | **0.676** | **0.737** |

Table 4 shows the comparison of the results with and without fine-tuning based on the COCO dataset. The synthetic images yielded good results in some classes, even outperforming training with equivalent original images (e.g., train). Some classes achieved good performance with just 10 reference images, while others, such as person, dog, and car, showed significant performance drops. Notably, the accuracy for the "person" class decreased significantly when using only reference images in synthetic-involved training sets. This may be due to the wide variation of person images in the COCO val2017 set (used for testing in this study), which the synthetic images could not fully capture. As a result, in the next section, we employed an advanced technique with more detailed control for the person class.

Table 4. The effect of finetuning.

| COCO (mAP@50) | car |
|---|---|
| Synthetic images (non-finetune) | 0.425 |
| Synthetic images (finetune) | **0.736** |

Given the original stable diffusion model capable of generating the daily classes of COCO, we compare the performance before and after the finetuning on a specific category of car. Table 4 shows the fine-tuning model easily outperforms the original model with more than a 30% improvement. This indicates the importance of fine-tuning to ensure the domain of the synthetic images is close to the target images' domain, which greatly benefits the final performance.

Table 5. The results of CottonWeedDet12.

| CottonWeedDet12 (mAP@50) | all | Purslane | Ragweed | PalmerAmaranth |
|---|---|---|---|---|
| original images | **0.99** | **0.99** | **1** | **0.986** |
| synthetic images | 0.983 | **0.99** | 0.98 | 0.97 |
| reference images | 0.696 | 0.78 | 0.567 | 0.843 |
| reference+synthetic images | 0.972 | 0.989 | 0.98 | 0.9 |

Table 5 shows the classification results for the CottonWeedDet12 dataset. The synthetic images, generated from just 10 reference images per weed species, achieved results comparable to those obtained from larger-scale original images. This demonstrates that advanced results can be achieved by capturing only a few realistic images and supplementing them with synthetic images generated through stable diffusion-based techniques.

### 3.3 ControlNet synthetic images

For COCO and CottonWeedDet12 datasets, three experiments were conducted for detection tasks. The first is the model performance of all the training images of the classes; the second is the model performance of sample images that were used to generate synthetic images; the third is the model performance of the synthetic images.

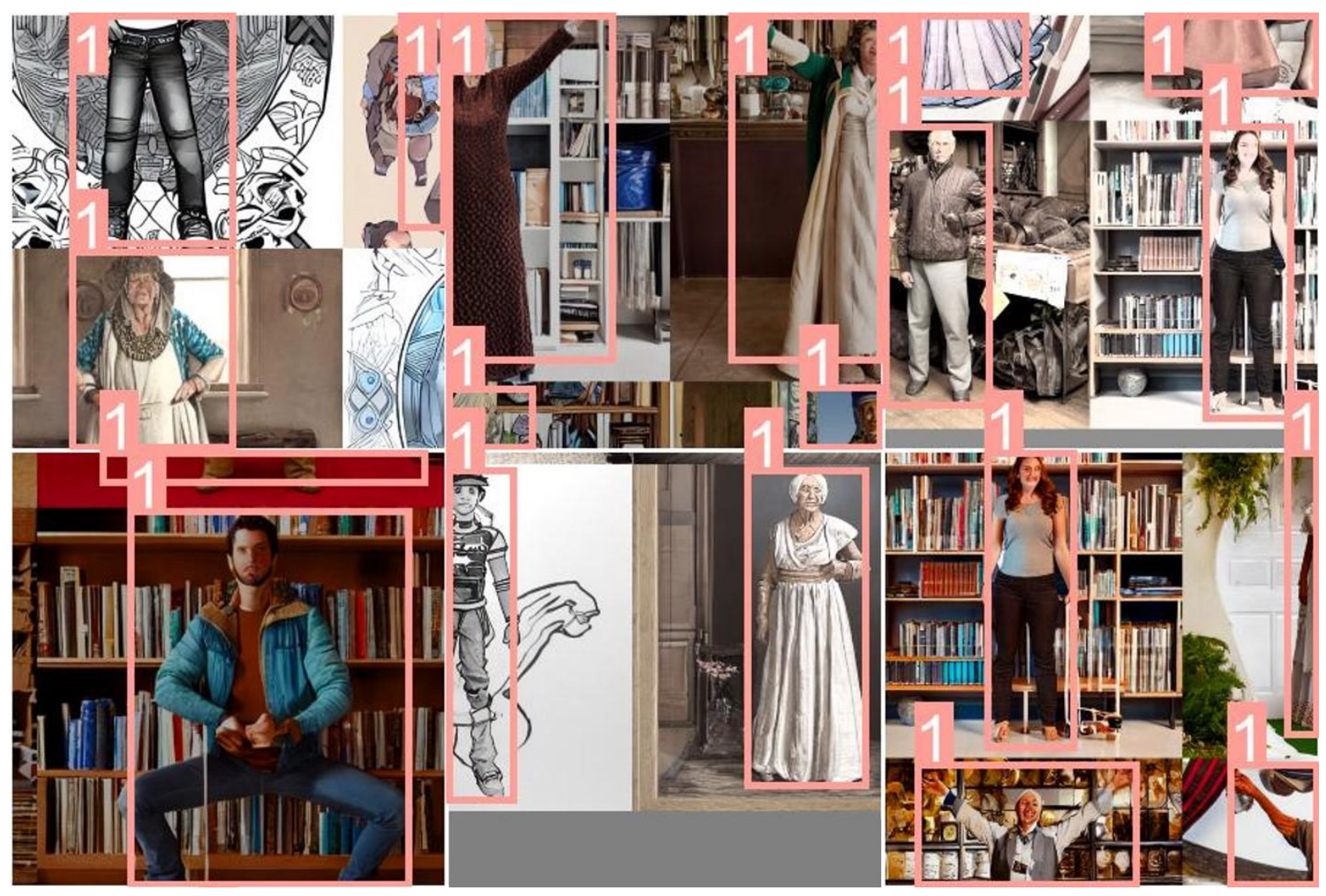

Figure 9. The preprocessed images of a batch size 8. Mosaic was used as the default preprocessor for YOLOv8.

Figure 9 shows the preprocessed images, and Table 6 compares the results trained on synthetic images to the original images. Despite the efforts of this study to generate a diversity of human images, the detection results of the synthetic-involved training set were worse than those only trained on the original images. Although somewhat frustrating, it is possible that only a few additional steps in this direction could surpass the original images, highlighting the need for further research.

Table 6. The detection results of the "person" class.

| COCO (mAP@50) | person |
|---|---|
| Original images | **0.842** |
| Synthetic images | 0.478 |
| Original+Synthetic images | 0.831 |

## Conclusion

Image-to-image translation can generate diverse variations of original images in both the COCO and CottonWeedDet12 datasets. DreamBooth allows for more specific image generation based on a small set of target images (e.g., daily objects or weed species), while ControlNet offers detailed control to diversify variations, such as the "person" class in the COCO dataset. In detection tasks, although synthetic images did not outperform original images in most cases, synthetic-included training sets yielded higher results in some classes. In classification tasks, classifiers trained on synthetic images generated from just 10 realistic reference images achieved results comparable to those trained on thousands of real images, marking a significant step forward in Stable Diffusion applications. Although this study only evaluated 7 classes in the COCO dataset and 3 weed species in CottonWeedDet12, the principles and insights here provide a strong foundation for incorporating stable diffusion-based synthetic images into both classification and detection tasks.